\documentclass[conference]{IEEEtran}
\IEEEoverridecommandlockouts
\usepackage{cite}
\usepackage{amsmath,amssymb,amsfonts}
\usepackage{algorithmic}
\usepackage{graphicx}
\usepackage{textcomp}
\usepackage{xcolor}
\usepackage{multirow}
\usepackage{float}
\usepackage{balance}
\usepackage{tikz}
\usetikzlibrary{shapes,decorations}

\newcommand{\shapelabel}[3]{
    \begin{tikzpicture}[baseline={([yshift=-0.8ex]current bounding box.center)}]
    \protect\node[#1,fill=#2,draw=black,scale=.8]{\color{white}\textbf{#3}};
    \end{tikzpicture}
}

\tikzset{
    shapelabel/.style n args={2}{
        #1,
        draw=black,
        fill=#2,
        baseline={([yshift=-0.8ex]current bounding box.center)},
        text=white
    }
}

\def\BibTeX{{\rm B\kern-.05em{\sc i\kern-.025em b}\kern-.08em
    T\kern-.1667em\lower.7ex\hbox{E}\kern-.125emX}}

\begin{document}

\title{Energy-Efficient Deployment of Machine Learning Workloads on Neuromorphic Hardware\\
}

\author{
    \IEEEauthorblockN{Peyton Chandarana, Mohammadreza Mohammadi, James Seekings, Ramtin Zand}
    \IEEEauthorblockA{Department of Computer Science and Engineering, University of South Carolina, Columbia, SC}
}

\makeatletter
\def\ps@IEEEtitlepagestyle{%
  \def\@oddfoot{\mycopyrightnotice}%
  \def\@evenfoot{}%
}
\def\mycopyrightnotice{%
  {\hfill \footnotesize 978-1-6654-6550-2/22/\$31.00 \copyright 2022 IEEE\hfill}
}
\makeatother

\maketitle
\begin{abstract} 
As the technology industry is moving towards implementing tasks such as natural language processing, path planning, image classification, and more on smaller edge computing devices, the demand for more efficient implementations of algorithms and hardware accelerators has become a significant area of research. In recent years, several edge deep learning hardware accelerators have been released that specifically focus on reducing the power and area consumed by deep neural networks (DNNs). On the other hand, spiking neural networks (SNNs) which operate on discrete time-series data, have been shown to achieve substantial power reductions over even the aforementioned edge DNN accelerators when deployed on specialized neuromorphic event-based/asynchronous hardware. While neuromorphic hardware has demonstrated great potential for accelerating deep learning tasks at the edge, the current space of algorithms and hardware is limited and still in rather early development. Thus, many hybrid approaches have been  proposed which aim to convert pre-trained DNNs into SNNs. 
In this work, we provide a general guide to converting pre-trained DNNs into SNNs while also presenting techniques to improve the deployment of converted SNNs on neuromorphic hardware with respect to latency, power, and energy. Our experimental results show that when compared against the Intel Neural Compute Stick 2, Intel's neuromorphic processor, Loihi, consumes up to 27$\times$ less power and 5$\times$ less energy in the tested image classification tasks by using our SNN improvement techniques.
\end{abstract}

\section{Introduction}
In the last 10 years, deep learning has transformed the technology industry enabling computers to perform image classification and recognition, translation, path planning, and more \cite{AKINOSHO2020101827, MISHRA2020104000, KOTSIOPOULOS2021100341, gpt2_fewshot}. While these efforts have been fruitful in terms of providing the desired functionality, most of these implementations employ the use of power-hungry hardware such as GPUs and TPUs \cite{googletpu} and are deployed in systems that are not constrained by power limitations. In recent, years many approaches have been proposed to alleviate these power constraints with methods such as quantization \cite{google_quantization, haq} and approximate computing \cite{2022_aptpu}. 
With these approaches,  many new edge-specific devices have been introduced such as the Nvidia Jetson Nano, Intel Neural Compute Stick 2, and Google Coral Edge TPU. Many of these edge devices were created to take advantage of quantized networks that operate on lower precision values rather than the standard single or double precision floating point representations. As a result, the overall power consumption and architecture area are reduced to specifically benefit applications where space and power are limited. 

Spiking neural networks (SNNs), considered the latest generation of artificial neural networks (ANNs), are a new class of neural networks that focus on biological plausibility, energy efficiency, and event-based computing \cite{2014_diehl_cook}. Unlike their continuous floating-point value-based deep neural network (DNN) counterparts, SNNs operate on discrete-temporal values which represent the biological action potentials of neurons in the brain \cite{2014_diehl_cook}. SNNs have been shown in many works \cite{heartbeat_loihi, image_seg_loihi, 2022_iopnce_mohammadi} to accomplish comparable accuracies to DNNs while also significantly reducing power and energy consumption. 



While, SNNs can be more energy and power efficient, 
training deep SNNs (DSNNs) has been a recurring challenge due to the lack of suitable training/learning algorithms that perform as well as the backpropagation algorithm used in DNNs \cite{2016_training_dsnn_backprop, 2018_rstdp, 2020_snn_training_dilemma}. 
Many SNN-specific learning algorithms have been proposed such as spike-timing dependent plasticity (STDP) and variants of it \cite{2018_rstdp, 2020_coding_selection}. These learning approaches rely on the temporal patterns found in the time between spikes to adapt the weight values as the network sees more input \cite{neuronal_dynamics}. This approach to learning, while efficient and suitable for SNNs of low depth dimensionality, do not typically scale well to deeper networks due to the lack of feedback from subsequent layers during training \cite{2016_training_dsnn_backprop, 2020_snn_training_dilemma}.
To address or even bypass the training and design challenges introduced in SNNs, many DNN to SNN conversion approaches have been proposed \cite{perez_conversion, cao_conversion, 2016_snntoolbox}. One such conversion approach, the SNN Conversion Toolbox \cite{2016_snntoolbox}, uses the parameters in pre-trained DNNs to create a similar SNN and deploy them on Loihi to provide energy-efficient and event-based computation to highly constrained environments in edge computing applications.
In this work, we aim to generalize the process of converting pre-trained DNNs into SNNs and deploying the SNN on neuromorphic hardware such as Loihi by contributing the following:
\begin{itemize}
    \item We provide general guidelines for designing and training DNNs for conversion into SNNs.
    \item After the SNNs are created, we present analysis and optimization techniques to further optimize the SNNs with respect to power, latency, and energy.
    \item We compare the performance of SNNs on Loihi against the Intel Neural Compute Stick 2 in classifying static images.
\end{itemize}

The remainder of this work is organized as follows. In Section \ref{sec:edgehardware}, we provide an overview of the two hardware platforms used in this work, the Intel Neural Compute Stick 2 and Intel Loihi, along with their respective APIs. In Section \ref{sec:conversion}, we discuss the conversion methodology and network considerations for converting DNNs to SNNs using the SNN Conversion Toolbox. We then provide some insights and techniques, in Section \ref{sec:deployment}, for optimizing the SNNs in terms of latency and energy consumption. In Section \ref{sec:results}, we present our experimental results with respect to inference accuracy, power, latency, and energy on the three separate image classification tasks. Finally, in Section \ref{sec:conclusion}, we conclude our work with a discussion of the findings and future directions of research.

\begin{figure}
    \centering
    \resizebox{.7\linewidth}{!}{
    \includegraphics[page=1]{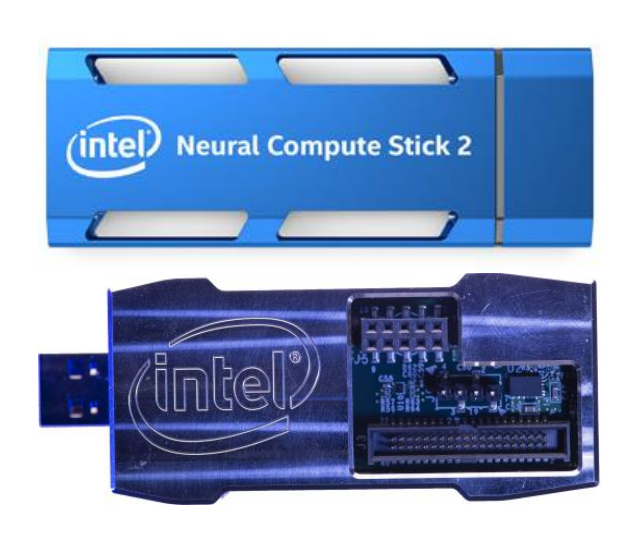}
    }
    \caption{Intel Neural Compute Stick 2 (top) \cite{ncs2_pic} and Intel Loihi Kapoho Bay (bottom) \cite{loihi_pics}.}
    \label{fig:kapoho_ncs2}
\end{figure}

\section{Edge Hardware}
\label{sec:edgehardware}
To demonstrate the benefits of neuromorphic hardware for machine learning tasks we use two hardware platforms along with their respective software APIs to perform our experiments. Here, we briefly describe the architectures and APIs of an edge computing neural network accelerator, the Intel Neural Compute Stick 2, and a neuromorphic hardware platform, Intel Loihi.
\begin{figure}
    \centering
    \resizebox{.6\linewidth}{!}{
    \includegraphics[page=2]{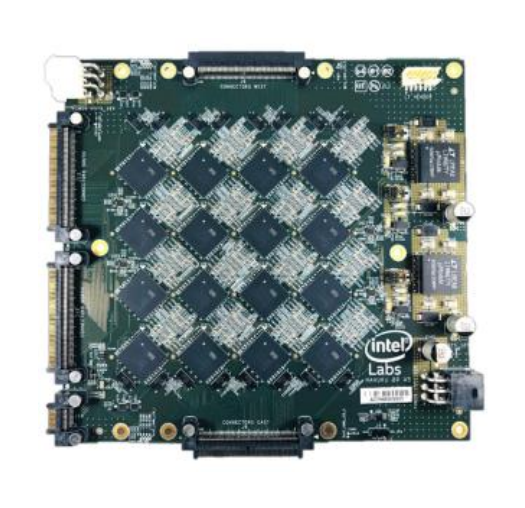}
    }
    \caption{Nahuku-32 Loihi server blade with 32 interconnected Loihi chips \cite{loihi_pics}.}
    \label{fig:nahuku32}
\end{figure}  
\subsection{Intel Neural Compute Stick 2}

In 2017, Intel launched the Movidius Neural Compute Stick meant to be used in edge computing devices to accelerate neural networks, specifically in computer vision-based applications using convolutional neural networks (CNNs). Since then, Intel has released an improved version called the Neural Compute Stick 2 (NCS2), which we use herein.

\begin{figure*}[h]
    \centering
    \includegraphics[width=\textwidth]{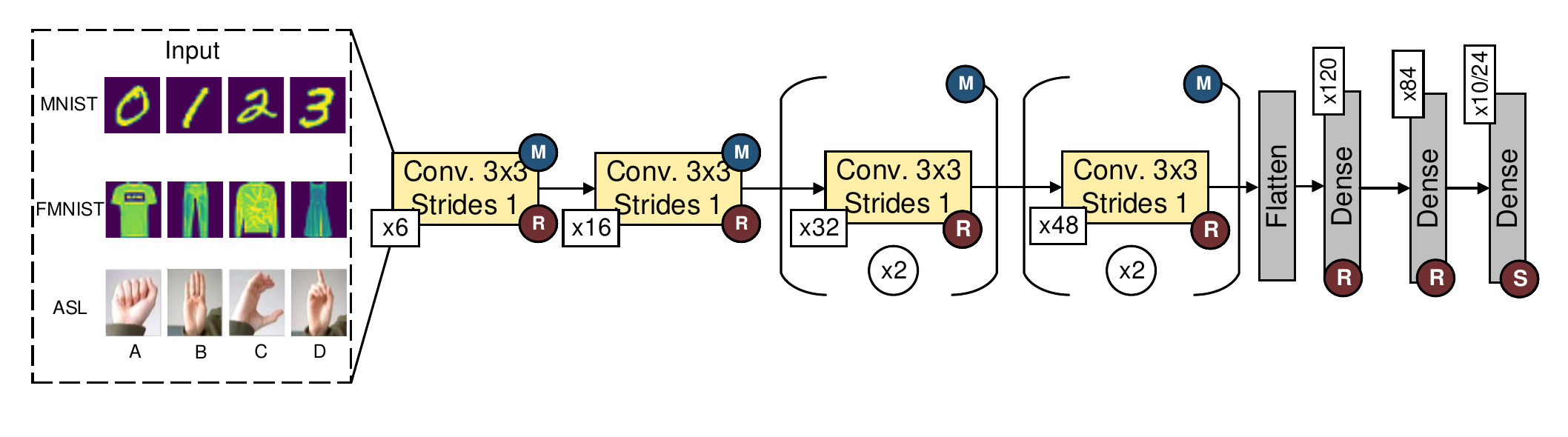}
    \caption{VGG-9 Network Architecture: \shapelabel{circle}{black!60!red}{R} ReLU, \shapelabel{circle}{blue!60!green}{M} MaxPooling2D, \shapelabel{circle}{blue!60!green}{D} Dropout, \shapelabel{circle}{black!60!red}{S} Softmax.}
    \label{fig:netarch}
\end{figure*}

The NCS2, shown in Fig. \ref{fig:kapoho_ncs2}, provides a plug-and-play USB interface for use with edge or small-computer devices like the Raspberry Pi. Specifically, NCS2 is clocked at 700 MHz and includes 16 shave cores, a neural engine, and 4 GB of memory which combine to implement a Vision Processing Unit (VPU) \cite{ncs2_specs}. Compared to conventional hardware's full or double-precision floating point operations, the Intel NCS2 only performs 16-bit floating-point operations allowing power and area savings at the cost of precision/accuracy.

To deploy models on NCS2, Intel provides an API called OpenVINO \cite{openvino} which allows models to be compiled and scaled/quantized for deployment on NCS2. Once the model is optimized for NCS2, the network can then be deployed on the device and input can be presented to perform inference.

\subsection{Intel Loihi}

Intel's neuromorphic platform, Loihi, was introduced in 2018 \cite{2018_loihi} with the goal of deploying SNNs on hardware to better establish neuromorphic computing's viability to accelerate tasks such as image classification and event-based or real-time computing problems. 

In the subsequent years since its launch in 2018, Loihi has been shown to have magnitudes lower energy consumption in machine learning applications while achieving comparable and even, in some cases, better accuracy than the traditional DNNs deployed on GPUs and TPUs. 
In terms of scalability, Loihi's hardware architecture enables it to be scaled from small USB form factor devices of one to four Loihi chips to the much larger data center implementations with many Loihi chips contained within a single server blade as seen in Figure \ref{fig:nahuku32} \cite{2018_loihi, 2021_loihi}. Each first-generation Loihi chip is comprised of 128 specialized-event-driven neuro-cores, each capable of implementing up to 1,024 spiking neurons in an SNN. Additionally, each neuro-core in the Loihi chip also contains 128 KB of state memory and allows the implementation of up to 4,096 fan-in or fan-out axons connecting to other neurons.

In this work, we employ the SNN Conversion Toolbox \cite{2016_snntoolbox} along with its custom Loihi backend, NxTF \cite{2021_nxtf}, to convert DNNs into SNNs to be deployed on Intel's Loihi platform. We go into further detail about this conversion process and methodology in Section \ref{sec:conversion}.




\section{DNN to SNN Conversion Methodology}
\label{sec:conversion}

Our experiments perform image classification on three distinct image datasets: the MNIST handwritten digit dataset \cite{mnist}, the fashion MNIST (FMNIST) clothing dataset \cite{fmnist}, and the American Sign Language (ASL) Alphabet \cite{asl_kaggle}. MNIST and FMNIST consist of 10 distinct classes each. MNIST contains 70,000 images in total of handwritten digits zero through nine. FMNIST, like MNIST, also contains 70,000 static images including images of different pieces of clothing such as pullovers, trousers, bags, etc. Unlike the MNIST and FMNIST, the ASL Alphabet dataset contains 24 classes representing static hand gestures corresponding to the English letters A thru Y, excluding the non-static gestures for letters J and Z. The ASL Alphabet dataset includes 34,627 static ASL gestures in total. Each of these datasets was split into the typical train, validation, and test subsets which are made up of 60\%, 20\%, and 20\% of the total images, respectively.

To eliminate any bias with respect to power, latency, and energy, we chose to use a modified version of the VGGNet network proposed in \cite{vggnet} for the representative DNN of our experiments. We use this network across the three datasets to ensure that the model performances discussed in Section \ref{sec:results} are not influenced by the network size or architecture. As seen in Figure \ref{fig:netarch}, our VGGNet implementation consists of six convolution layers each using the ReLU activation function and followed by a max-pooling layer. After these six convolution layers, we flatten the output and then pass the resulting features into three dense layers consisting of 120, 84, and either 10 or 24 output neurons depending on the input dataset. In total, this implementation of VGGNet, which we call VGG-9 for the remainder of this work, has 66,378 or 67,568 parameters for the MNIST/FMNIST and the ASL Alphabet datasets, respectively.

After training these VGG-9 networks, we use SNN Conversion Toolbox \cite{2016_snntoolbox} for converting our DNNs to more energy-efficient SNNs.
In \cite{2016_snntoolbox}, Rueckauer et. al. use the weights and activations of pre-trained DNNs to map DNN layers and neurons to the spiking domain in a one-to-one manner. By mapping these layers and neurons 
to construct the SNN, the challenges of training and designing SNNs can be somewhat avoided. 

The SNN Conversion Toolbox, compared to the previous conversion works \cite{perez_conversion, cao_conversion}, implements most of the common layers used in DNNs like convolution layers, pooling layers, and activation functions. However, there are some limitations as to what type of pooling layers and activation functions can be converted into spiking equivalents. For example, while the SNN Conversion Toolbox does implement max pooling layers for its built-in simulator, the NxTF backend \cite{2021_nxtf} for Loihi does not due to max-pooling needing special implementation considerations at the neuron level. The SNN Conversion Toolbox also does not support the hyperbolic tangent or TanH activation function. For these reasons, we constrain our implementation of the VGG-9 model to employ average-pooling layers as opposed to the classical VGGNet's max-pooling layers. We refer to these constrained networks in Section \ref{sec:results} as C-DNNs. As we will later see in Section \ref{sec:results}, this change does not significantly impact the network's accuracy and in some cases can even improve the network's accuracy.

After training our constrained VGG-9 networks, we then use the SNN Conversion Toolbox to convert the network into an SNN to be deployed on Loihi. The first steps performed by the SNN Conversion Toolbox consist of normalizing the DNN parameters with respect to SNNs, mapping the neurons/layers to spiking equivalents, and converting the input data into sequences of spikes or spike trains. To convert the input data into spike trains the SNN Conversion Toolbox uses a rate-based coding approach to encode an input image's pixel intensities into spike trains that have a proportional spike rate, called the firing rate, to the pixel intensity \cite{2016_snntoolbox}. That is, the higher the pixel intensity, the higher the spike frequency of a pixel's corresponding spike train. These spike trains are then exposed to the SNN for a fixed user-configurable parameter called \textit{duration} measured in milliseconds. In our experiments, we run each SNN for numerous iterations with different durations. This duration parameter directly affects the latency of the network as seen in \cite{2022_iopnce_mohammadi}. Similar to \cite{2022_iopnce_mohammadi}, our analysis includes searching for a duration that reduces the duration parameter and therefore latency with little impact on accuracy.

\begin{table}
\centering
\caption{Model Parameters and Loihi Core Partitioning}
\resizebox{.8\linewidth}{!}{
\begin{tabular}{lrc}
\hline \\[-8pt]
Layer & Parameter & Loihi Cores \\ \hline  \hline \\[-8pt]
Conv1 & 60 & 12 \\
Conv2 & 880 & 7 \\
Conv3 & 4640 & 4 \\
Conv4 & 9248 & 7 \\
Conv5 & 13872 & 2 \\
Conv6 & 20784 & 3 \\
FC1 & 5880 & 1 \\
FC2 & 10164 & 1 \\
Output & 850\,\textbf{/}\,2040 & 1 \\ \hline \\[-8pt]
Total & 66,378\,\textbf{/}\,67,568 & 47 \\ \hline
\end{tabular}
}
\label{table:netparams}
\end{table}

\section{Deployment Methodology}
\label{sec:deployment}
While the SNN Conversion Toolbox did not originally support deployment on Loihi, a custom backend for the SNN Conversion Toolbox was released in 2021, called NxTF \cite{2021_nxtf}. This backend was built using Intel's Loihi API called NxSDK and implemented the spiking layers the conversion toolbox was capable of converting. With this layer/neuron implementation, NxTF also included a custom hardware partitioning/distribution optimization algorithm to efficiently distribute the SNN's layers and neurons across the neuro-cores as well as across multiple Loihi chips if needed. In the case of our VGG-9 networks, only one Loihi chip was used by the deployed SNNs. In Table \ref{table:netparams}, we show the layer-by-layer parameters and the number of neuro-cores that the layers were allocated. Once we begin inference on the deployed SNN on Loihi, we perform two types of optimization to (1) reduce latency while maximizing accuracy and (2) to increase data sparseness to reduce overall SNN power and energy consumption.

\begin{figure}[]
    \centering
    \resizebox{.9\linewidth}{!}{
    \includegraphics[]{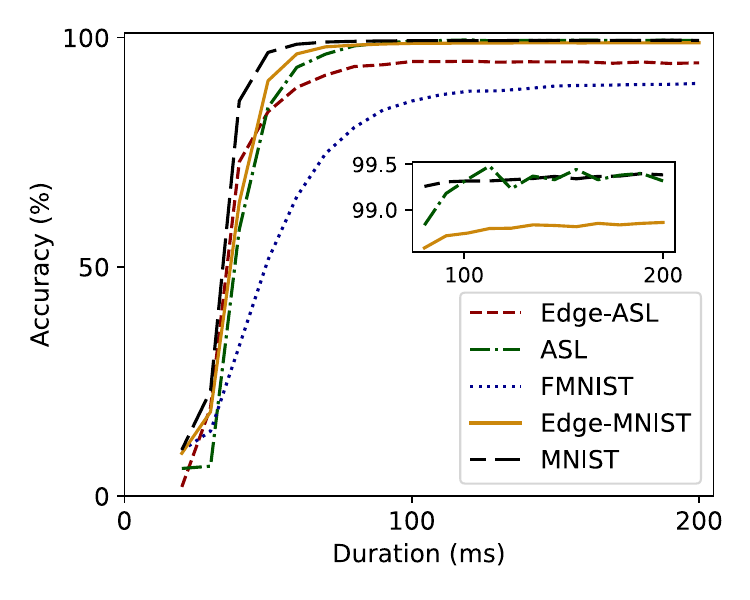}
    }
    \caption{Accuracy vs. Duration with zoomed portion above inflection point.}
    \label{fig:accuracy_vs_duration}
\end{figure}

One challenge we encountered while performing our experiments was the polling rates of the power and energy consumption hardware sensors.
In this first iteration of Loihi, these sensors operate on timescales of 30 to 40 ms \cite{ncl_models}. Thus, this hardware polling limitation constrains the range of durations for which power and energy metrics can be recorded. To alleviate this issue, we ran multiple durations above the 30-40 ms polling threshold and then averaged the total power consumption. From there, we ran the remaining durations below the polling threshold without measuring power and energy consumption. From our experiments, there is a linear correlation between duration and latency. Using this fact, we are able to predict an upper limit of energy consumption by using a predicted latency value along with the average power consumption measured above the polling threshold.

\begin{figure}[h]
    \centering
    \resizebox{.6\linewidth}{!}{
    \includegraphics{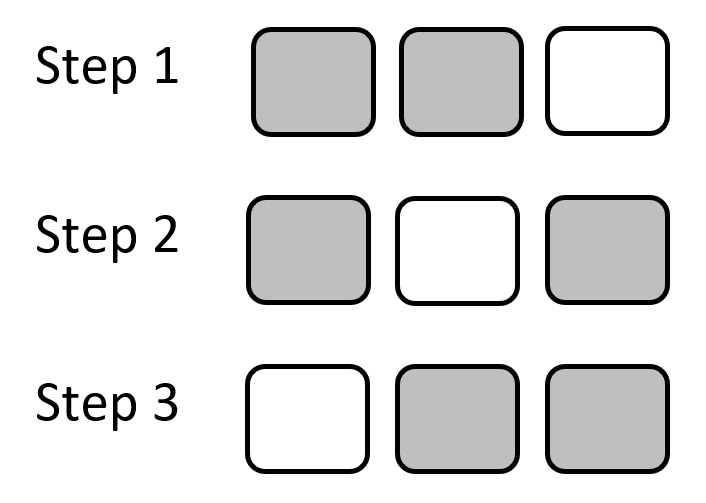}
    }
    \caption{Test dataset subsets used in the proposed latency/duration cross-validation process.}
    \label{fig:crossvalidation}
\end{figure}

\subsection{Duration/Latency Cross-Validation}
After all layers in the SNN are partitioned to corresponding neuro-cores, the SNN is deployed using a user-configurable parameter file which includes the aforementioned duration parameter. Throughout our experiments, we change this duration parameter to optimize the network's latency and accuracy. As shown in Figure \ref{fig:accuracy_vs_duration}, the longer the input spike trains are presented to the network the more accurate inference becomes. However, as can also be seen in Figure \ref{fig:accuracy_vs_duration}, there is an inflection point for which the accuracy of all networks plateau after increasing the duration above a specific point. Thus, increasing duration above this point causes the latency of the network to increase significantly with little to no gain in accuracy seen in the zoomed portion of Figure \ref{fig:accuracy_vs_duration}. Herein, we present a method to optimize the network to minimize the duration/latency while maximizing the accuracy within 2.0\% of the maximum accuracy.

To optimize the duration without overfitting to a specific test dataset during inference, we use an approach that is similar to a cross-validation approach typically employed in DNN training. To start, we first performed inference on our networks for durations ranging from 10 ms to 200 ms in 10 ms intervals using a 2/3 subset of our test dataset. After collecting the durations along with their corresponding accuracies, we then searched this accuracy-duration space for an optimal point that minimizes the duration while maximizing the accuracy within 2.0\% of the network's maximum SNN accuracy. With this optimal point found, we then used the remaining 1/3 subset of the test dataset to test the duration point to ensure reasonable accuracy was still attainable. We then repeated this process two additional times with two different dataset splitting configurations. Figure \ref{fig:crossvalidation} shows how we split the dataset into two for finding the optimal point and then testing it on a separate dataset.



\subsection{Increasing Sparsness with Edge Detection}

\begin{figure}
    \centering
    \resizebox{.8\linewidth}{!}{
    \includegraphics{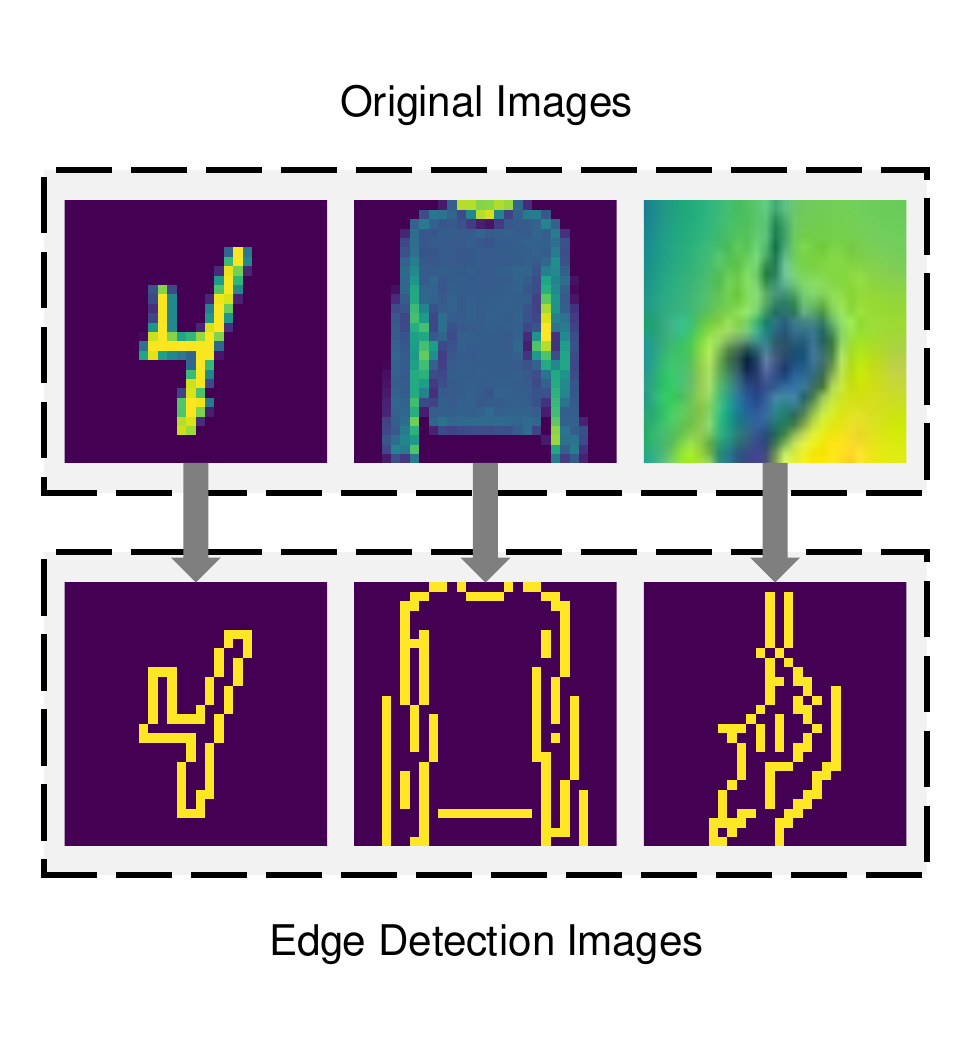}
    }
    \caption{Original images converted to edge images using Canny edge detection.}
    \label{fig:edgeconversion}
\end{figure}

In many works \cite{2018_sparse_computation_asnn, 2021_backprop_sparse_reg, 2022_sparse_compressed_snn}, it has been shown that introducing sparseness into SNNs can achieve further reductions in power, latency, and energy. 
In \cite{2021_igsc_chandarana}, using edge detection is shown to significantly reduce the input data and therefore should reduce the power and energy consumption.
Thus, to increase the sparseness of the input data, we first perform a preprocessing step that applies OpenCV's \cite{opencv_library} Canny edge detection on the inputs prior to inference. By applying Canny edge detection the input data can be reduced and thus lower the spike rate/count by only allowing the neurons of the edge pixels to receive input. As shown in Figure \ref{fig:edgeconversion}, the edge detected images are much more sparse and binary compared to their original images. These edge-detected images are then input into the SNN and the same experiments are performed and the power, latency, and energy are recorded.

\section{Results}
\label{sec:results}
Here we present our experimental results which consist of comparing the DNN and SNN implementations of our VGG-9 networks on the MNIST, FMNIST, and ASL image classification tasks. We first compare the DNNs to the SNNs in terms of accuracy using both the original images and edge-detected images. In the case of FMNIST, we have omitted the edge detection results as reasonable accuracies were not attainable using the same VGG-9 network. We then compare the performance of NCS2 and Loihi in terms of optimal duration/latency, accuracy, power, and energy.

\subsection{Inference Accuracy}

\begin{table}
\centering
\caption{Maximum Accuracy of models trained on regular images vs. edge detected images}
\label{table:accuracy}
\resizebox{\linewidth}{!}{
\begin{tabular}{clccclccc}
\hline \\[-8pt] 
\multirow{2}{*}{Dataset} &  & \multicolumn{3}{c}{Regular Images} &  & \multicolumn{3}{c}{Edge Images} \\ \cline{3-5} \cline{7-9}  \\[-8pt] 
 &  & DNN & C-DNN & SNN &  & DNN & C-DNN & SNN \\ \hline \\[-8pt]
ASL &  & \textbf{99.83} & 99.50 & 99.54 &  & \textbf{96.90} & 95.47 & 94.90 \\
MNIST &  & 99.28 & 99.38 & \textbf{99.79} &  & 98.84 & 98.92 & \textbf{99.04} \\
FMNIST &  & \textbf{91.00} & 90.13 & 89.71 &  & - & - & - \\ \hline
\end{tabular}
}
\end{table}

\begin{table*}[h]
\centering
\caption{NCS2 vs. Loihi Power and Latency for Original Images and Edge Images Before Duration/Latency Optimization}
\label{table:power}
\resizebox{\linewidth}{!}{
\begin{tabular}{lccccccccccc}
\hline
\multicolumn{1}{c}{\multirow{3}{*}{Dataset}} & \multicolumn{3}{c}{NCS2} & \multicolumn{1}{l}{} & \multicolumn{7}{c}{Loihi} \\ \cline{2-4} \cline{6-12} 
\multicolumn{1}{c}{} & \multicolumn{2}{c}{\begin{tabular}[c]{@{}c@{}}Power \\ (mW)\end{tabular}} & \multirow{2}{*}{\begin{tabular}[c]{@{}c@{}}Latency\\ (ms)\end{tabular}} & \multicolumn{1}{l}{} & \multicolumn{3}{c}{\begin{tabular}[c]{@{}c@{}}x86 Cores Power \\ (mW)\end{tabular}} & \multicolumn{3}{c}{\begin{tabular}[c]{@{}c@{}}Neuro-Cores Power\\ (mW)\end{tabular}} & \multirow{2}{*}{\begin{tabular}[c]{@{}c@{}}Latency\\ (ms)\end{tabular}} \\ \cline{2-3} \cline{6-11}
\multicolumn{1}{c}{} & Idle & Running &  & \multicolumn{1}{l}{} & Static & Dynamic & Total & Static & Dynamic & Total &  \\ \hline
ASL & 635 & 1465 & 2.27 &  & 0.136 & 19.05 & 19.19 & 21.54 & 14.00 & 35.54 & 11.56 \\
MNIST & 635 & 1470 & 2.21 &  & 0.136 & 19.75 & 19.89 & 21.52 & 22.15 & 43.67 & 10.10 \\
FMNIST & 635 & 1472 & 2.20 &  & 0.134 & 19.22 & 19.35 & 21.25 & 26.16 & 47.41 & 12.25 \\ \hline
Edge-ASL & 635 & 1438 & 2.50 &  & 0.136 & 19.00 & 19.14 & 21.57 & 9.79 & 31.36 & 10.85 \\
Edge-MNIST & 635 & 1453 & 2.25 &  & 0.136 & 19.40 & 19.54 & 21.60 & 12.31 & 33.91 & 11.24 \\ \hline
\end{tabular}
}
\end{table*}


In Table \ref{table:accuracy}, the DNNs, with the exception of FMNIST, performed rather well in both the original image and edge detected image classification tasks achieving as high as 99.83\% and 98.92\% accuracy, respectively. The constrained or C-DNN for MNIST actually achieves better accuracy than the original DNN in both the original and edge-detected cases. The accuracies for FMNIST are the lowest of the three datasets running on our VGG-9 network achieving 91.00\% on the DNN and 90.13\% on the C-DNN.

When comparing the DNNs and C-DNNs to the converted SNNs, Table \ref{table:accuracy} shows that the MNIST SNNs actually outperform both the DNN and C-DNN for both the original images and edge images with a respective 0.41\% and 0.12\% increase in accuracy. With the exception of FMNIST, it appears that even with most of the information being removed by edge detection, the MNIST and ASL models are still able to achieve reasonable accuracies at 99.04\% and 96.90\%, respectively.

\subsection{Power, Latency, and Energy}

In Tables \ref{table:power} and \ref{table:comp} we provide the power, latency, and energy that yield the maximum attainable accuracy in our experiments. 
As seen in Table \ref{table:power} the power metrics for both the NCS2 and Loihi remain somewhat similar between the different dataset experiments. This behavior is expected as the networks are very similar in architecture only differing in the output layer.
Comparing Loihi and NCS2 in terms of total running power, Table \ref{table:power} shows that Loihi consumes around $\sim22\times$ less power than NCS2 under load. 

While the original image SNNs already realize significant power reductions, Table \ref{table:power} shows that performing edge detection can further reduce power consumption by as much as $\sim44.42\%$ compared to SNNs without edge detection. Thus, the SNNs on Loihi with edge detection are even more efficient consuming approximately $\sim27\times$ less total power than NCS2's best-case total power consumption. This improvement in power consumption results from Loihi's asynchronous/event-based computations, only consuming additional power when non-zero stimuli are presented. This is in contrast to DNNs where even though input data may be sparse, MAC operations are still performed independent of the input data and thus consume more power.

In Table \ref{table:comp} we provide the accuracy, inference power, latency, and inference energy metrics for the optimal duration/latencies. 
While there is a small decrease in accuracy, $<2.0\%$, due to our latency/duration optimization when comparing Tables \ref{table:accuracy} and \ref{table:comp}, the latency is reduced by at least $\sim9.77\%$ and up to $\sim37.63\%$ shown in Tables \ref{table:power} and \ref{table:comp}. Since power measurements are not possible on Loihi below the aforementioned 30-40 ms polling of the hardware sensors, we have calculated the energy for these optimized latencies by multiplying the expected total power from Table \ref{table:power} by the optimal latencies. According to Table \ref{table:comp} Loihi has worse latency than NCS2. However, this may improve with future iterations of Loihi as \cite{2021_loihi} describes a significant overhead introduced due to the x86 and neuro-core communications. Even though the latency is higher on Loihi, Table \ref{table:comp} shows that the energy consumed by Loihi is $\sim3.08\times$ to $\sim5.00\times$ lower than NCS2.



\begin{table}[]
\centering
\caption{NCS2 vs Loihi - Post-Latency Optimization Comprehensive Analysis}
\label{table:comp}
\begin{tabular}{clcccc}
\hline
\multirow{2}{*}{Hardware} & \multicolumn{1}{c}{\multirow{2}{*}{Dataset}} & \multicolumn{4}{c}{Benchmarking Metrics} \\ \cline{3-6} 
 & \multicolumn{1}{c}{} & \begin{tabular}[c]{@{}c@{}}Accuracy\\ (\%)\end{tabular} & \begin{tabular}[c]{@{}c@{}}Power\\ (mW)\end{tabular} & \begin{tabular}[c]{@{}c@{}}Latency\\ (mS)\end{tabular} & \begin{tabular}[c]{@{}c@{}}Energy\\ (mJ)\end{tabular} \\ \hline
\multirow{5}{*}{NCS2} & ASL & 99.83 & 830.00 & 2.27 & 1.88 \\
 & MNIST & 99.28 & 835.00 & 2.21 & 1.85 \\
 & FMNIST & 91.00 & 837.00 & 2.20 & 1.85 \\ \cline{2-6} 
 & Edge-ASL & 96.90 & 803.00 & 2.50 & 2.01 \\
 & Edge-MNIST & 98.84 & 818.00 & 2.25 & 1.85 \\ \hline
\multirow{5}{*}{Loihi} & ASL & 98.19 & 54.73 & 9.34 & 0.51 \\
 & MNIST & 98.55 & 63.56 & 6.13 & 0.39 \\
 & FMNIST & 88.25 & 66.76 & 9.03 & 0.60 \\ \cline{2-6}
 & Edge-ASL & 93.73 & 50.50 & 9.79 & 0.49 \\
 & Edge-MNIST & 97.91 & 53.45 & 7.01 & 0.37 \\ \hline
\end{tabular}
\end{table}
\section{Conclusion}
\label{sec:conclusion}

Herein, we presented methods to constrain and train deep neural networks for conversion into spiking neural networks. In addition to providing some general network design considerations, we proposed two techniques that aim to optimize the latency and power consumption of these SNNs when deployed on neuromorphic hardware. The first technique is based on using a cross-validation method during inference to decrease latency. The other method involved using a pre-processing method such as edge detection to add sparsity to the input of the SNN models to further decrease inference power. Our results, exhibited that Intel's Loihi neuromorphic processor achieves similar, if not better, accuracy to Intel's Neural Computer Stick 2 DNN accelerator while reducing energy consumption by a factor of 5$\times$ when using our proposed energy-efficient deployment strategies. 

\section*{Acknowledgment}
\noindent
This work is partially supported by an ASPIRE grant from the Office of the Vice President for Research at the University of South Carolina. Special thanks to the Intel Neuromorphic Research Community (INRC) for providing access to the Loihi chips for the experiments performed in this paper.
\balance
\bibliographystyle{IEEEtran}
\bibliography{refs}

\end{document}